\documentclass[review]{elsarticle}

\usepackage{lineno,hyperref}

\journal{Pattern Recognition}

\newcommand{\ie}{\textit{i.e.}}

\newcommand{\etc}{\textit{etc}}

\usepackage{amsmath,amssymb,amsfonts}
\usepackage{algorithmic}
\usepackage{graphicx}
\usepackage{textcomp}
\usepackage{subfigure}
\usepackage{multirow}
\usepackage{multicol}
\usepackage{comment}
\usepackage{ulem}
\usepackage{soul}
\usepackage{tabularx}
\usepackage{tabu}
\usepackage{booktabs}
\usepackage{makecell}
\usepackage{enumerate}
\usepackage{url}
\usepackage{algorithm}
\DeclareMathOperator*{\argmax}{argmax}

\usepackage{arydshln}
\usepackage{wrapfig}

\begin{document}

\begin{frontmatter}

\title{CrossRectify: Leveraging Disagreement for Semi-supervised Object Detection}


\author[CASIA,UCAS1]{Chengcheng Ma}
\author[Youtu]{Xingjia Pan}
\author[UCAS2]{Qixiang Ye}
\author[Jilin]{Fan Tang\corref{mycorrespondingauthor}}
\ead{tangfan@jlu.edu.cn}
\author[CASIA]{Weiming Dong}
\author[CASIA]{Changsheng Xu}

\address[CASIA]{National Lab of Pattern Recognition (NLPR), Institute of Automation, Chinese Academy of Sciences (CASIA), Beijing, 100190, China}
\address[UCAS1]{School of Artificial Intelligence, University of Chinese Academy of Sciences (UCAS), Beijing, 100049, China}
\address[Youtu]{Youtu Lab, Tencent Inc., Shanghai, 200233, China}
\address[UCAS2]{School of Electronic, Electrical and Communication Engineering, University of Chinese Academy of Sciences (UCAS), Beijing, 101408, China}
\address[Jilin]{Jilin University, Changchun, 130000, China}

\cortext[mycorrespondingauthor]{Corresponding author}
\vspace{-2mm}
\begin{abstract}
Semi-supervised object detection has recently achieved substantial progress. 
As a mainstream solution, the self-labeling-based methods train the detector on both labeled data and unlabeled data with pseudo labels predicted by the detector itself, but their performances are always limited.
Through experimental analysis, we reveal the underlying reason is that the detector is misguided by the incorrect pseudo labels predicted by itself (dubbed self-errors).
These self-errors can hurt performance even worse than random-errors, and can be neither discerned nor rectified during the self-labeling process.
%
In this paper, we propose an effective detection framework named CrossRectify, to obtain accurate pseudo labels by simultaneously training two detectors with different initial parameters.
Specifically, the proposed approach leverages the disagreements between detectors to discern the self-errors and refines the pseudo label quality by the proposed cross-rectifying mechanism.
%
Extensive experiments show that CrossRectify achieves outperforming performances over various detector structures on 2D and 3D detection benchmarks.
\end{abstract}

\end{frontmatter}


\section{Introduction}
\label{sec:intro}
The success of deep learning has greatly promoted the development of object detection approaches~\cite{SSD,FasterRCNN,bosquet2021stdnet,wang2021multi,kong2021spatial,zhang2021cadn}, and a large amount of labeled data is essential to the training process of object detectors.
However, as illustrated in \cite{kuznetsova2020open}, it is always labor-intensive and expensive to acquire a large amount of labeled data with bounding-box-level annotations. 
In comparison, unlabeled data are much easier and cheaper to collect.
Therefore, semi-supervised object detection~\cite{SSOD} is recently investigated to reduce the cost of data annotating, which leverages only a few labeled data and a large amount of unlabeled data to train object detectors.

\begin{figure}[t]
\centering
{
\includegraphics[width=.72\linewidth]{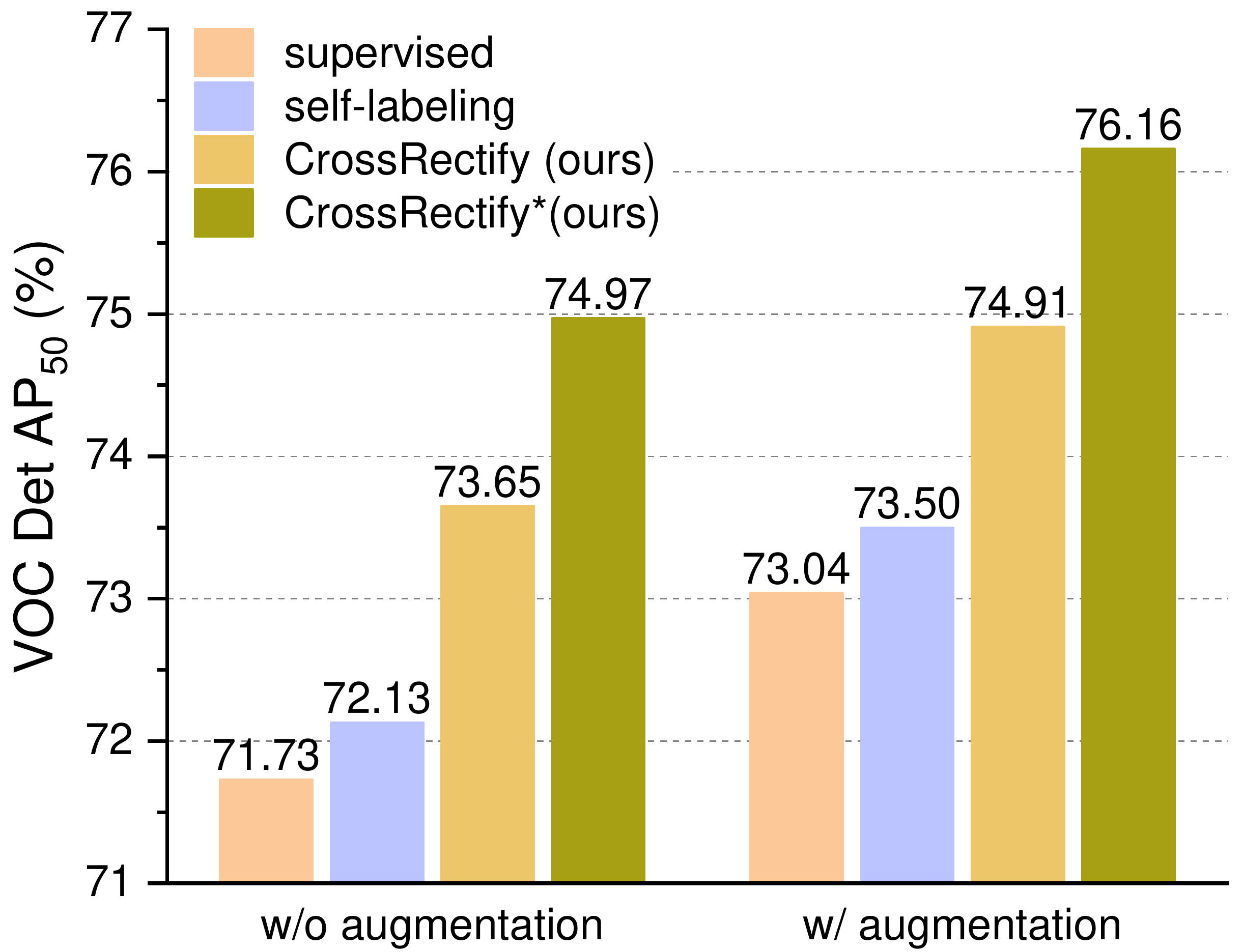}
}
\caption{The proposed CrossRectify and CrossRectify$^\ast$ method can outperform the self-labeling-based semi-supervised object detection method by large margins on the Pascal VOC benchmark dataset without/with the mix-up data augmentation.
(For interpretation of the references to color in this figure legend, please refer to the online version of this article.)
} 

\label{fig:first-page}
\end{figure}

Semi-supervised object detection (SSOD) has achieved significant progress in recent years, and one mainstream of existing SSOD solutions is based on the self-labeling scheme~\cite{self-labeling}.
The core idea of self-labeling scheme is to first utilize the current detector to predict pseudo bounding boxes for unlabeled data in each training iteration, then conduct detector training with both labeled data and pseudo-labeled data. 
However, compared with fully-supervised baselines, the performance increments brought by the self-labeling-based methods are always observed limited. 
For example, the absolute AP$_{50}$ gain is only 0.40\% (0.46\%) without (with) the mix-up data augmentation~\cite{MixUp} over the SSD300~\cite{SSD} detector on the Pascal VOC~\cite{VOC} benchmark (as shown in Fig.~\ref{fig:first-page}).
To reveal the reason behind such phenomenon, we introduce the ground-truth annotations of unlabeled data for in-depth analysis.
Existing self-labeling-based SSOD methods generate pseudo labels by selecting bounding boxes with high confidence scores to ensure the quality of pseudo labels.
However, we observe that part of high-confidence boxes are still misclassified (dubbed self-errors), and these self-errors can hurt the detection performance even worse.
Specifically, when we replace the misclassified category labels with random labels (dubbed random-errors) during the self-labeling training process, the final performance is even improved.
%
Unfortunately, these self-errors can be neither \textit{discerned} nor \textit{rectified} by the detector itself, which we summarize as two inherent limitations of the self-labeling training scheme.
%
These two limitations will lead the detector to be misguided by self-errors, and  finally results in insignificant performance increments.
%

Since one single detector can neither discern nor rectify the misclassified pseudo bounding boxes, it is necessary to introduce the guidance information from other distinct detector to the training process.
Recently, a few works \cite{CoTraining,ke2019dual,wei2020combating,coteaching+} illustrate the fact that two differently initialized models with an identical structure can yield diverse results on the same training sample during the training process. 
Inspired by this fact, in this paper, we propose an effective and general training framework named CrossRectifiy for both 2D and 3D semi-supervised object detection task.
%
In CrossRectify, two detectors with the same structure but different initialization are trained simultaneously, and each detector is supervised by the pseudo labels generated from the proposed cross-rectifying mechanism.
Specifically, the proposed cross-rectifying mechanism first leverages the disagreements on the same objects between two detectors to discern the latent self-errors predicted by each single detector. 
%
Then, the pseudo labels are generated based on the bounding boxes predicted by two detectors, by adopting a simple yet effective comparison-to-assigning pipeline with confidence scores being considered.
%
To this end, the proposed CrossRectify method can discern and rectify the self-errors and improve the pseudo label quality.

%
Note that a few recent works utilize another separate detector for pseudo label generation, which can somehow alleviate the limitations of self-labeling training scheme. 
For instance, \cite{STAC,DataUncertainty,UnbiasedTeacher,HumbleTeacher,SoftTeacher} adopt the teacher-student mutual learning framework~\cite{MeanTeacher} and utilize a specific teacher detector to generate pseudo labels in an offline or online way.
However, these approaches still suffer from the following shortcomings:
the pseudo labels always remain fixed~\cite{STAC,DataUncertainty}, and the teacher detector is converged to the student detector in the late stage of training~\cite{UnbiasedTeacher,HumbleTeacher,SoftTeacher}, thus the labeling process degenerates into the self-labeling manner and suffers from the same limitations.
Similar with our method, \cite{InstantTeaching} proposes the co-rectify method to train two models simultaneously and takes the average on two predictions sets as pseudo labels, which is the only prior work adopting the co-training framework~\cite{CoTraining,ke2019dual,wei2020combating,coteaching+} to the best of our knowledge.
However, we conduct quantitative comparison on the pseudo label quality, and the experimental results validate the superiority of our CrossRectify method compared with co-rectify.

We carry out extensive experiments on both 2D and 3D semi-supervised object detection tasks to verify the effectiveness and versatility of the proposed CrossRectify method. 
As illustrated in Section \ref{sec: results}, our method obtains consistent and substantial improvements compared with the state-of-the-art SSOD methods on the Pascal VOC, MS-COCO, and SUN-RGBD benchmark datasets, improving by $>$1\% absolute AP margins.

Our main contributions are summarized as follows:
\begin{enumerate}[1)]
\item We point out that the performances of self-labeling-based SSOD approaches are always limited, and the reason behind such phenomenon lies in that the detector can neither discern nor rectify the misclassified pseudo bounding boxes predicted by itself.
\item We propose an effective approach named CrossRectify to discern and rectify the misclassified pseudo bounding boxes using the disagreements between two detectors, which can address the inherent limitations of self-labeling and improve the detection performance.
\item We conduct extensive experiments on both 2D and 3D object detection benchmark datasets, and the results verify the superiority of the proposed CrossRectify approach, compared with state-of-the-art approaches.
\end{enumerate}

In the remainder of this paper, we briefly review several related works in Section~\ref{sec:rw}. 
Then we describe the two limitations of self-labeling-based SSOD approaches in Sec~\ref{sec:methodology}, and provide the technical details of the proposed CrossRectify method in Sec~\ref{sec:CrossRectify}.
Finally, we display the detection performances of 2D and 3D SSOD tasks in Sec~\ref{sec:exp} and conclude the whole paper in Sec~\ref{sec: conclusion}.

\section{Related Work}
\label{sec:rw}

\subsection{2D and 3D object detection}
\label{sec:rw1}
Object detection is one of the most significant tasks in computer vision, including 2D and 3D scenes. 
In the field of 2D object detection, the structures of detectors can be categorized into single-stage (SSD~\cite{SSD}, \etc) and two-stage (Faster-RCNN~\cite{FasterRCNN}, cascade RCNN~\cite{cai2018cascade}, \etc), depending on whether a region proposal network is utilized. 
Although these detectors have reached outstanding performances, their training processes heavily rely on a large amount of labeled data with bounding box annotations, which are always laborious and expensive to acquire~\cite{kuznetsova2020open}.
In this paper, we focus on how to leverage a large amount of unlabeled data for performance increment. 
For fair comparisons with existing works, we conduct experiments over the SSD300 and Faster-RCNN-FPN detector structures on 2D object detection, and over
the VoteNet~\cite{votenet} structure on 3D object detection.

\subsection{Semi-supervised Object Detection (SSOD)}
\label{sec:rw2}
\noindent The existing SSOD approaches can be categorized as follows.
\paragraph{Consistency regularization} Many of the existing SSOD methods utilize the consistency regularization proposed in semi-supervised learning (SSL), such as CSD~\cite{CSD}, ISD~\cite{ISD}, PL~\cite{PL}, \etc. 
The key idea of consistency regularization is to require the detector to predict consistently on both weak- and strong- augmented versions of the same input. 
%
We point out that the consistency-regularization-based methods can be regarded as the special case of the self-labeling training scheme, because the detector is supervised by the pseudo labels on weak-augmented image predicted by itself.
Note that these studies report large performance improvements over the fully-supervised baselines, but we point out that the experimental settings are somehow unfair. 
In fact, under the same data augmentation, the performance increments brought by consistency regularization are always observed limited.
We analyze the reason behind such phenomenon in this paper, and point out the inherent limitations of the self-labeling training scheme, as well as the consistency regularization.
%

\paragraph{Teacher-student mutual learning} Beyond the consistency regularization, there are a few SSOD works based on the teacher-student mutual learning framework~\cite{MeanTeacher}. 
In an offline or online manner, the pseudo labels are generated by the teacher model, instead of by the student model itself. 
As for the former, \cite{STAC,DataUncertainty} first pre-train the teacher model with available labeled data, then utilize the teacher to annotate the entire unlabeled data. 
However, the pseudo labels are generated only once and remain fixed during semi-supervised training, thus the final performance of student model is limited by that of the teacher model.
As for the latter, ~\cite{UnbiasedTeacher,HumbleTeacher,SoftTeacher} compute the exponential moving average of student as teacher and the teacher is utilized to generate pseudo labels during the semi-supervised training process. 
However, we observe that the teacher is converged to the student in the late stage of training, which indicates that the labeling process degenerates into the self-labeling manner and suffers from the same limitations.

\paragraph{Co-rectify and CPS}
As another research line of semi-supervised learning, co-training methods~\cite{CoTraining,ke2019dual,wei2020combating,coteaching+} are proposed to train two models in a collaborative manner. 
Each model can learn from the pseudo labels predicted by its counterpart, which seems a promising way to mitigate the limitations of self-labeling methods. 
The only prior work taking the idea of co-training for the SSOD task is co-rectify~\cite{InstantTeaching}. In~\cite{InstantTeaching}, the pseudo boxes for unlabeled data are first predicted by one detector, then refined by the corresponding predictions from another model, with probability scores and coordinates being averaged. 
%
Besides, a recent study~\cite{CPS} also adopts co-training and proposes the cross pseudo supervision (CPS) for the semi-supervised semantic segmentation task, where each model is supervised by the pseudo maps predicted by the other model. 
We note that CPS can be adapted for the SSOD task.
However, we conduct quantitative comparisons to show that both co-rectify and CPS cannot fully exploit the advantages of multiple models and improve the quality of pseudo labels.

\section{Problem Analysis}
\label{sec:methodology}

In this section, we first introduce the preliminaries in semi-supervised object detection (SSOD), then analyze the inherent limitations of self-labeling-based SSOD methods.

\subsection{Preliminaries}
\label{sec:preliminaries}
Under the semi-supervised setting, an object detector $f$ is trained on a labeled dataset $D_l= \{\boldsymbol x_i^l, \boldsymbol y_i^l\}_{i=1}^{N_l}$ with $N_l$ samples and an unlabeled dataset $D_u= \{\boldsymbol x_j^u\}_{j=1}^{N_u}$ with $N_u$ samples.
For a labeled image $\boldsymbol x^l$, its annotation $\boldsymbol y^l=(\boldsymbol c, \boldsymbol t)$ contains the category labels $\boldsymbol c$ and coordinates $\boldsymbol t$ of all foreground objects.

Overall, the detector model $f$ is optimized by minimizing the supervised loss $L_S$ on labeled data and unsupervised loss $L_U$ on unlabeled data, formulated as:
\begin{flalign}
L = L_{S} + \lambda_{U} \cdot L_{U},
\label{eq: loss_total}
\end{flalign}
where $\lambda_{U}$ denotes the weight factor.
Generally, the supervised loss $L_S$ consists of the classification loss $l_{cls}$ and coordinate regression loss $l_{reg}$:
\begin{flalign}
L_S = l_{cls}\big(f_{cls}(\boldsymbol x^l), \boldsymbol c \big)+l_{reg}\big(f_{loc}(\boldsymbol x^l), \boldsymbol t \big),
\label{eq: supervised loss}
\end{flalign}
where $f_{cls}(\boldsymbol x^l)$ and $f_{loc}(\boldsymbol x^l)$ stand for probabilities and coordinates predicted by the classification and localization branch of detector $f$, respectively.

In each training iteration, the self-labeling-based SSOD method utilizes the current detector to predict bounding boxes on the unlabeled inputs $\boldsymbol x^u$, and then selects the pseudo labels $\hat{\boldsymbol y}$ with confidence larger than the threshold $\tau$, and finally computes the unsupervised loss:
\begin{flalign}
L_U = l_{cls}\big(f_{cls}(\boldsymbol x^u), \hat{\boldsymbol c}\big)
    + l_{reg}\big(f_{loc}(\boldsymbol x^u), \hat{\boldsymbol t}\big),
\label{eq: self-labeling}
\end{flalign}
where $\hat{\boldsymbol y}=(\hat{\boldsymbol c}, \hat{\boldsymbol t})=\big(\argmax f_{cls}(\boldsymbol x^u), f_{loc}(\boldsymbol x^u)\big)$
and $\max f_{cls}(\boldsymbol x^u) > \tau$ are correspondingly satisfied.


As the special case of self-labeling training, the consistency-regularization-based methods introduce weak data augmentation $\alpha(\cdot)$ and strong data augmentations $A(\cdot)$, and the Eq.~(\ref{eq: self-labeling}) is updated as:
\begin{flalign}
L_U = l_{cls}\big(f_{cls}(A(\boldsymbol x^u)),\argmax f_{cls}(\alpha(\boldsymbol x^u)) \big) 
    + l_{reg}\big(f_{loc}(A(\boldsymbol x^u)), f_{loc}(\alpha(\boldsymbol x^u)) \big).
\label{eq: consistency regularization}
\end{flalign}
Note that as for the consistency regularization, the loss term Eq.~(\ref{eq: consistency regularization}) can be computed on both labeled and unlabeled data, and the strong augmentations can also boost the performances of fully-supervised training~\cite{ISD}.
Accordingly, the total loss in Eq.~(\ref{eq: loss_total}) is augmented as $L = L_{S} + \lambda_{U} \cdot \big(L_U(\boldsymbol x^l) + L_U(\boldsymbol x^u)\big)$, and the fully-supervised baseline is trained by optimizing $L = L_{S} + \lambda_{U} \cdot L_U(\boldsymbol x^l)$.

\begin{table}[t]
\centering
\renewcommand\arraystretch{1.0}
\caption{Results of self-labeling-based semi-supervised object detection methods under various data augmentations. 
The benchmark dataset is Pascal VOC and the detector structure is SSD300.
``SeLa'' stands for self-labeling. 
``identical'', ``HF'' and ``MU'' stand for no data augmentation, horizontal flip augmentation (CSD)~\cite{CSD} and mix-up augmentation (ISD)~\cite{ISD}, respectively. 
The figures in brackets are the performance increments over the fully-supervised baselines, which always seem trivial for
the self-labeling-based methods.}
\vspace{2.5mm}
\scalebox{0.75}{
\begin{tabular}{ l c c c | l }
\hline
\makecell[c]{Method} & $A(\cdot)$ & Labeled & Unlabeled & \makecell[c]{$\text{AP}_{50}$ (\%)}
\\ \hline
Supervised       & identical               & VOC07 &   -    & 71.73
\\ 
SeLa             & identical               & VOC07 & VOC12  & 72.13 (+0.40)
\\
Supervised       & identical               & VOC0712 &  -   & 77.37 (+5.64)
\\ \hline
Supervised       & HF                      & VOC07 &   -    & 71.89
\\
SeLa             & HF                      & VOC07 & VOC12  & 72.35 (+0.46)
\\ 
Supervised       & HF                      & VOC0712 &  -   & 77.26 (+5.37)
\\ \hline
Supervised       & MU                      & VOC07 &   -    & 73.04
\\ 
SeLa             & MU                      & VOC07 & VOC12  & 73.50 (+0.46)
\\
Supervised       & MU                      & VOC0712 &  -   & 78.83 (+5.79)
\\\hline
\end{tabular}
}
\label{tab: self-labeling-based-performances}
\end{table}

\subsection{Limitations of Self-labeling}
\label{sec:over-fitting in sela}
To verify the performances of existing self-labeling-based SSOD methods, we conduct experiments on the Pascal VOC benchmark dataset~\cite{VOC} based on the SSD300 structure~\cite{SSD}. 
We use the trainval set of VOC07 as labeled data and trainval set of VOC12 as unlabeled data, and finally report the AP$_{50}$ performance on the test set of VOC07. 
The confidence threshold $\tau$ is fixed as 0.5, similar with \cite{CSD} and \cite{ISD}.
For comparison, we conduct fully-supervised training with same hyper-parameters (batch size, iteration number, \etc) as baseline. 
As shown in Table~\ref{tab: self-labeling-based-performances}, the AP$_{50}$ improvement achieved by self-labeling methods is only 0.40\%.
Besides, we test with two representative consistency-regularization-based methods, namely CSD~\cite{CSD} and ISD~\cite{ISD}. 
%
Similarly, the detection results achieve only 0.46\% absolute AP$_{50}$ gains over the baseline, which indicates the inefficiency of self-labeling training scheme.

\begin{table}[t]
\centering
\scriptsize
\renewcommand\arraystretch{0.62}
\caption{Analysis on the limitations of self-labeling methods. 
The benchmark dataset is Pascal VOC and the detector structure is SSD300. 
TP (FP) stands for the correctly (falsely) classified pseudo boxes. 
``SeLa'' stands for self-labeling, which uses both TP and FP for training. 
The figures in brackets are the AP$_{50}$ increments over the fully-supervised baseline.}
\vspace{2.5mm}
\begin{tabular}{ l | c | c | c | l }
\hline
\makecell[c]{Method} & Labeled & Unlabeled & $\tau$ & \makecell[c]{$\text{AP}_{50}$}
\\ \hline
Supervised            & VOC07   & -      & -                   & 71.73 
\\ 
SeLa                  & VOC07   & VOC12  & 0.5                 & 72.13 (+0.40)
\\ 
SeLa                  & VOC07   & VOC12  & 0.5$\rightarrow$0.8 & 72.09 (+0.36)
\\ 
SeLa                  & VOC07   & VOC12  & 0.8                 & 72.12 (+0.39)
\\ \hline
SeLa (use TP and discard FP)    & VOC07   & VOC12  & 0.5                 & 74.03 (+2.30)
\\ 
SeLa (use TP and random labeled FP)& VOC07   & VOC12  & 0.5                 & 73.87 (+2.14)
\\
SeLa (use GT labels for TP and FP) & VOC07   & VOC12  & 0.5                 & 74.86 (+3.13)
\\ 
Supervised            & VOC0712 & -      & -                   & 77.37 (+5.64)
\\ \hline
\end{tabular}
\label{tab: overfitting-analysis}
\vspace{1mm}
\end{table}
\begin{figure}[t]
\centering
\scalebox{0.95}
{
\includegraphics[width=1\linewidth]{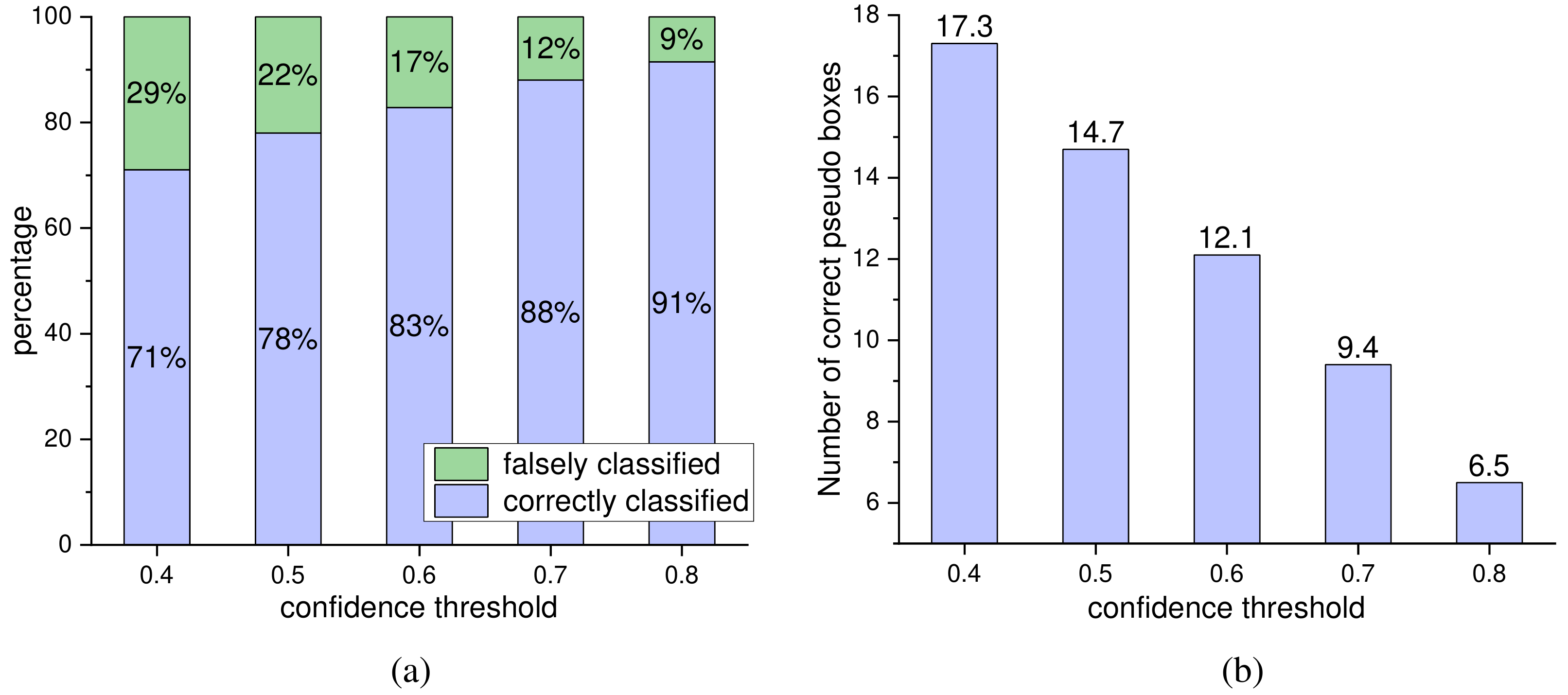}
}
\vspace{-3.mm}
  \caption{Pseudo label quality under different confidence thresholds in self-labeling training. (a) Precision of pseudo labels. (b) Average number of correctly classified pseudo boxes in each iteration. (For interpretation of the references to color in this figure legend, please refer to the online version of this article.)}
  \label{fig:confidence-threshold}
  \vspace{-2.mm}
\end{figure}

To reveal the reason behind such phenomenon and find out possible solutions, we introduce the ground-truth category labels of all pseudo bounding boxes for in-depth analysis.
Although all existing self-labeling-based SSOD methods generate pseudo labels by selecting bounding boxes with high confidence scores to ensure the quality of pseudo labels, Fig.~\ref{fig:confidence-threshold}(a) illustrates that part of high-confidence pseudo bounding boxes can also be misclassified. 
We name these incorrect boxes ``self-errors'' for clarity.
Since all pseudo boxes are predicted by the detector, it is impossible for the detector itself to discern the self-errors, which we summarize as the inherent limitation of the self-labeling process. 
We further conduct an experiment to illustrate how much can such limitation affect the detection performance:
in each training iteration, when we use the correct pseudo bounding boxes and discard the incorrect ones for training, the AP$_{50}$ result increases from 72.13\% to 74.03\% (see the 2nd and 5th rows in Table~\ref{tab: overfitting-analysis}).
Besides, we note that the detection performance cannot be improved by naively increasing the confidence threshold $\tau$.
As shown in Fig.~\ref{fig:confidence-threshold}(a) and (b), the precision of pseudo bounding boxes increases from 71\% to 91\% as the threshold increases from 0.4 to 0.8,
while a large threshold can also overkill the correct pseudo boxes and waste the unlabeled training data. 
Correspondingly, we conduct self-labeling training under three settings of the threshold $\tau$: (i) fixed as 0.5, (ii) fixed as 0.8, (iii) rising from 0.5 to 0.8 gradually during the whole training process. 
As displayed in Table~\ref{tab: overfitting-analysis} (from 2nd row to 4th row), the trade-off between precision and recall leads to similar final performances (about 72.1\% AP$_{50}$).
%

Since the self-errors cannot be discerned during the self-labeling process, it is also impossible for the detector to rectify them, which we summarize as the second limitation of the self-labeling process. 
In each training iteration, when we utilize the ground-truth (GT) category labels for all pseudo bounding boxes, the AP$_{50}$ result increases to 74.86\%, obtaining a 3.13\% absolute gain rather than 0.40\% (see the penultimate row in Table~\ref{tab: overfitting-analysis}). 
Such phenomenon shows the effectiveness of pseudo label rectification.
Besides, we find another interesting phenomenon: 
when we replace the misclassified category labels with random labels (dubbed random-errors) during the self-labeling training process, the final performance increases to 73.87\% (see the third row from the bottom in Table~\ref{tab: overfitting-analysis}). 
We also conduct experiments on the Faster-RCNN-FPN structure~\cite{FasterRCNN} and observe the similar trend: the AP$_{50}$ obtains with a 0.2\% gain by replacing the misclassified labels with random labels during training.
Such phenomena imply that the detector model can be misguided more severely by self-errors than random-errors. 

Based on the above experimental analysis, we draw that two inherent limitations in self-labeling-based SSOD methods will lead the detector to be misguided by self-errors, and self-errors will hurt the detector performance even worse than random-errors.

\section{Methodology}
\label{sec:CrossRectify}
\subsection{CrossRectify}
Since one single detector can neither discern nor rectify its self-errors, an intuitive idea is to utilize another model to deal with them.
%
Inspiring by the fact that two models with same structure but different initialization can yield different predictions on the same input~\cite{CoTraining,ke2019dual,wei2020combating,coteaching+}, we present the \textbf{CrossRectify} method to address the inherent limitations of self-labeling.
%

\begin{figure*}[t]
\centering
\vspace{-5.0mm}
\scalebox{0.92}
{
\includegraphics[width=1\linewidth]{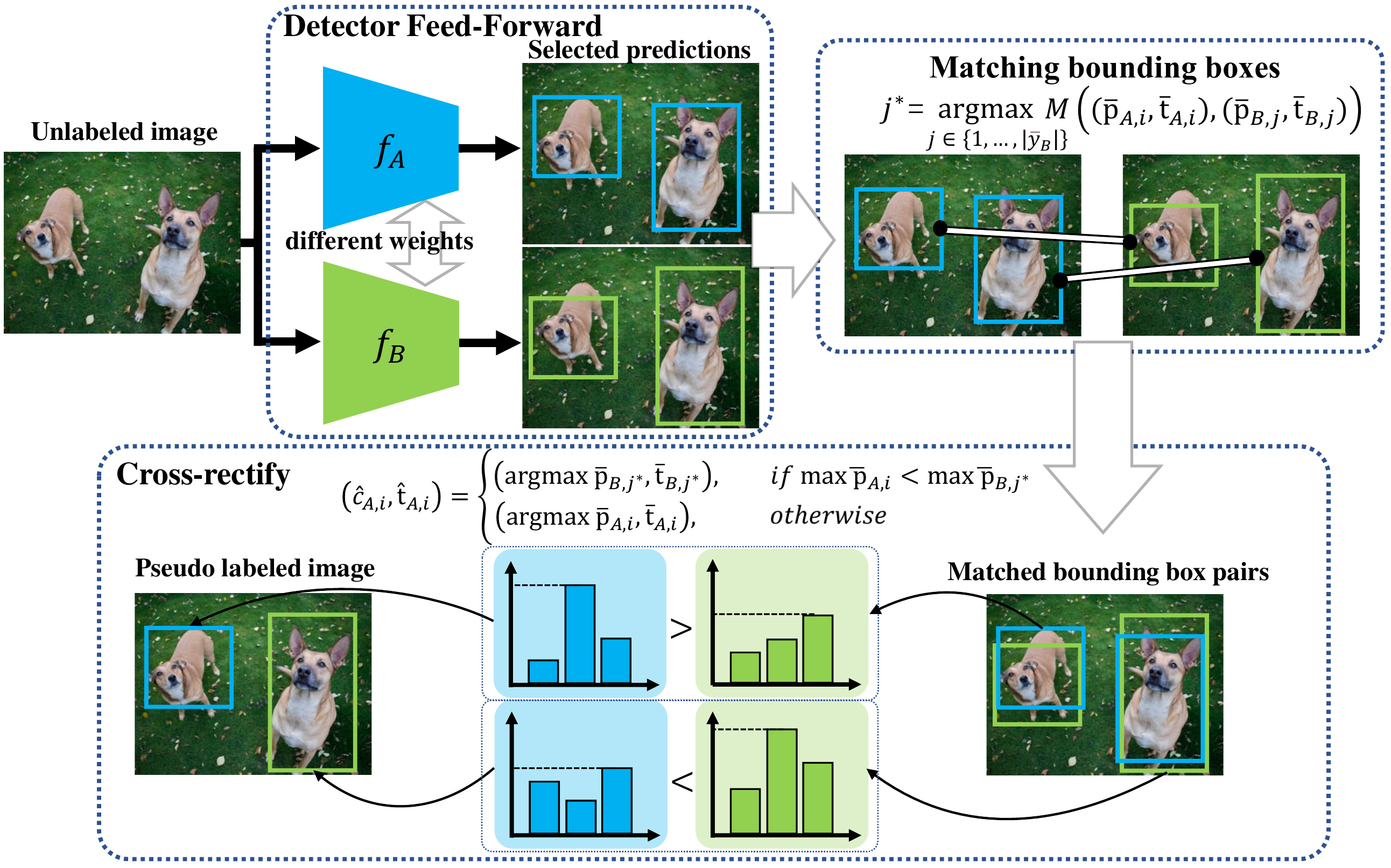}
}
\vspace{-0.mm}
  \caption{Overview of the pseudo label generation process in the proposed algorithm. Refer to Section\ref{sec:CrossRectify} for more details.
  (For color discrimination in this figure, please refer to the online version of this article.)}
  \label{fig:overview}
  \vspace{-0.mm}
\end{figure*}

In CrossRectify, two detectors with the same structure but distinct initialization\footnote{Take 2D object detection task for example. The backbone parameters are both initialized by the ImageNet-pretrained model, while the parameters of detection heads are randomly initialized.}, $f_A$ and $f_B$, are trained simultaneously. 
Both detectors are trained by jointly optimizing the supervised and unsupervised loss in Eq.~(\ref{eq: loss_total}).
For simplicity, we only introduce how to generate pseudo bounding boxes $\hat{\boldsymbol y}_A$ for training detector $f_A$, since the pseudo boxes $\hat{\boldsymbol y}_B$ for training detector $f_B$ are generated in the same way.
There are three steps in generating $\hat{\boldsymbol y}_A$, including: 
{\bf 1)} conducting detector feed-forward;
{\bf 2)} matching predicted bounding boxes;
{\bf 3)} cross-rectifying the matched boxes to generate pseudo labels.
They will be explained in details sequentially.
The label generation process of $\hat{\boldsymbol y}_A$ is illustrated in Fig.~\ref{fig:overview} and briefly summarized in Algorithm~\ref{alg: cross rectify}.

\paragraph{Detector feed-forward}\
In each training iteration, we utilize two detector models $f_A$ and $f_B$ to predict on the unlabeled inputs, then select the bounding boxes with max probability scores higher than threshold $\tau$, denoted as $\overline{\boldsymbol y}_A=(\overline{\boldsymbol p}_A, \overline{\boldsymbol t}_A)$ and $\overline{\boldsymbol y}_B=(\overline{\boldsymbol p}_B, \overline{\boldsymbol t}_B)$. $\overline{\boldsymbol p}$ and $\overline{\boldsymbol t}$ denote the probability scores and coordinates of all predicted bounding boxes.

\paragraph{Matching bounding boxes}\
For each box in $\overline{\boldsymbol y}_A$, we search its best match box among all boxes in $\overline{\boldsymbol y}_B$. 
For example, for \textit{i}-th box $(\overline{\mathbf{p}}_{A,i},\overline{\mathbf{t}}_{A,i})$ in $\overline{\boldsymbol y}_A$, the matching process is formulated as:
\begin{flalign}
j^\ast = \argmax_{j\in\{1,\cdots,\left| \overline{\boldsymbol  y}_B \right|\}} M\Big((\overline{\mathbf{p}}_{A,i},\overline{\mathbf{t}}_{A,i}),(\overline{\mathbf{p}}_{B,j},\overline{\mathbf{t}}_{B,j})\Big),
\label{eq: best-match}
\end{flalign}
where $M(\cdot,\cdot)$ stands for the matching metric and is slightly different for kinds of detectors. 
For the Faster-RCNN~\cite{FasterRCNN}, $M((\overline{\mathbf{p}}_{A,i},\overline{\mathbf{t}}_{A,i}),(\overline{\mathbf{p}}_{B,j},\overline{\mathbf{t}}_{B,j}))$ is the area of intersection over union (IoU) between two boxes as:
\begin{flalign}
M\big((\overline{\mathbf{p}}_{A,i},\overline{\mathbf{t}}_{A,i}),(\overline{\mathbf{p}}_{B,j},\overline{\mathbf{t}}_{B,j})\big)
=
\frac{\overline{\mathbf{t}}_{A,i} \cap \overline{\mathbf{t}}_{B,j}}
{\overline{\mathbf{t}}_{A,i} \cup \overline{\mathbf{t}}_{B,j}}.
\label{eq: iou}
\end{flalign}
Specifically, if the IoU areas between  $(\overline{\mathbf{p}}_{A,i},\overline{\mathbf{t}}_{A,i})$ and all boxes in $\overline{\boldsymbol y}_B$ are all below a certain threshold $\delta$, we create a virtual bounding box $\overline{\mathbf{t}}_{B,j^\ast}$ to match it, as $(\overline{\mathbf{p}}_{B,j^\ast},\overline{\mathbf{t}}_{j^\ast})=(\overline{\mathbf{p}}_{A,i},\overline{\mathbf{t}}_{A,i})$ and the matching metric $M(\cdot,\cdot)$ equals 1.
For the SSD structure~\cite{SSD}, $M((\overline{\mathbf{p}}_{A,i},\overline{\mathbf{t}}_{A,i}),(\overline{\mathbf{p}}_{B,j},\overline{\mathbf{t}}_{B,j}))$ equals 1 if two boxes are based on the same anchor, otherwise 0.
Note that $M(\cdot,\cdot)$ for SSD can also be specified as the IoU areas as like that for Faster-RCNN, but we find it more effective to adopt the anchor correspondence in our experiments.
For the VoteNet~\cite{votenet}, the matching metric $M(\cdot,\cdot)$ is specified as the negative Euclidean distance between the centers of two bounding boxes, computed as:
\begin{flalign}
M\big((\overline{\mathbf{p}}_{A,i},\overline{\mathbf{t}}_{A,i}),(\overline{\mathbf{p}}_{B,j},\overline{\mathbf{t}}_{B,j})\big)
=
-\left \| C(\overline{\mathbf{t}}_{A,i}) - C(\overline{\mathbf{t}}_{B,j}) \right \|_2,
\label{eq: votenet matching metric}
\end{flalign}

\noindent where $C(\cdot)$ denotes the center of a certain box.



\begin{algorithm}[t]
\caption{Generating pseudo bounding boxes $\hat{\boldsymbol y}_A$ via CrossRectify for training detector $f_A$.}
\label{alg: cross rectify}

\begin{algorithmic}[1]
\REQUIRE Object detectors $f_A$ and $f_B$, and unlabeled input. 
\ENSURE  The pseudo bounding boxes for training $f_A$.
\STATE Utilize $f_A$ and $f_B$ to predict on unlabeled input
\STATE Select the predicted boxes with their max probability scores larger than threshold $\tau$, denoted as $\overline{\boldsymbol y}_A$ and $\overline{\boldsymbol y}_B$.
\STATE Compute the matching metrics $M(\cdot,\cdot)$ for each bounding boxes in $\overline{\boldsymbol y}_A$ with all bounding boxes in $\overline{\boldsymbol y}_B$.
\STATE Find the best matching box for each bounding boxes in $\overline{\boldsymbol y}_A$ (using Eq.~\ref{eq: best-match}).
\STATE Compare max probability scores between each matched pair and obtain the pseudo boxes (using Eq.~\ref{eq: core equation})
\end{algorithmic}

\end{algorithm}

\paragraph{Cross-rectifying}
Based on the matched bounding box pair, each pseudo bounding box $(\hat{c}_{A,i}, \hat{\mathbf{t}}_{A,i})$ within $\hat{\boldsymbol y}_A$ can be generated as: 
\begin{equation}
\label{eq: core equation}
(\hat{c}_{A,i}, \hat{\mathbf{t}}_{A,i})=\begin{cases}   
(\argmax\overline{\mathbf{p}}_{B,j^\ast},\overline{\mathbf{t}}_{B,j^\ast}), & \text{\textit{if} $\max \overline{\mathbf{p}}_{A,i} < \max \overline{\mathbf{p}}_{B,j^\ast}$}\\   
(\argmax\overline{\mathbf{p}}_{A,i},\overline{\mathbf{t}}_{A,i}), & \text{\textit{otherwise}}.    
  \end{cases}
\end{equation}
\noindent Note that the Eq.~(\ref{eq: core equation}) covers two situations.
(a) When both detectors predict the same class on a certain object, we adopt it as the pseudo label, since two decisions are more reliable than one decision. 
(b) When two detectors have disagreements on a certain object, such bounding box tends to be unreliable.
To this end, the bounding box with higher confidence is regarded as the pseudo label.
The rationality behind such cross-rectifying mechanism lies in that the bounding boxes with higher confidence scores are more likely to be correctly classified (see Fig.~\ref{fig:confidence-threshold}(a)).
Thus, the wrong elements in $\overline{\boldsymbol y}_A$ can be both discerned and rectified in such manner.

When the training process ends, we evaluate the performance of one single detector. 
Moreover, to exploit the different detection abilities, we propose to adopt the weighted boxes fusion (WBF)~\cite{WBF} strategy to ensemble two predictions sets. The corresponding performance is denoted as \textbf{CrossRectify$^\ast$}.

\subsection{Comparisons with other works}
\label{sec: discussions}
Now we conduct quantitative analysis to show the superiority of CrossRectify on improving pseudo label quality, comparing with other recent works.  
 
\paragraph{Teacher-student mutual learning}
As discussed in Section \ref{sec:rw2}, some recent SSOD works~\cite{STAC,DataUncertainty,UnbiasedTeacher,HumbleTeacher,SoftTeacher} are established on the offline/online teacher-student mutual learning~\cite{MeanTeacher}. 
These works can alleviate the self-errors in self-labeling process by introducing another separate object detector for generation of pseudo labels.
However, as for offline methods~\cite{STAC,DataUncertainty}, the pseudo labels are generated only once and remain fixed when training student detector, so the student performance is upper-bounded by that of teacher.
For instance, we conduct experiments with Faster-RCNN-FPN detector on the MS-COCO benchmark dataset under 10\% degree of supervision.
The AP$_{50}$ performance of teacher detector after fully-supervised pre-training is 23.86\%, while that of student detector supervised by teacher detector only increases with a 3.30\% absolute gain, far away from the results in Table~\ref{tab: MS-COCO results} (34.89\% AP$_{50}$). 
Similar phenomenon can be observed with respect to the SSD300 structure on the Pascal VOC benchmark dataset in Table~\ref{tab: VOC results} (obtaining only a 0.79\% AP$_{50}$ gain).

\begin{figure}[t]
\centering
{
\begin{minipage}[t]{0.49\linewidth}
\centering
\includegraphics[width=0.9\linewidth]{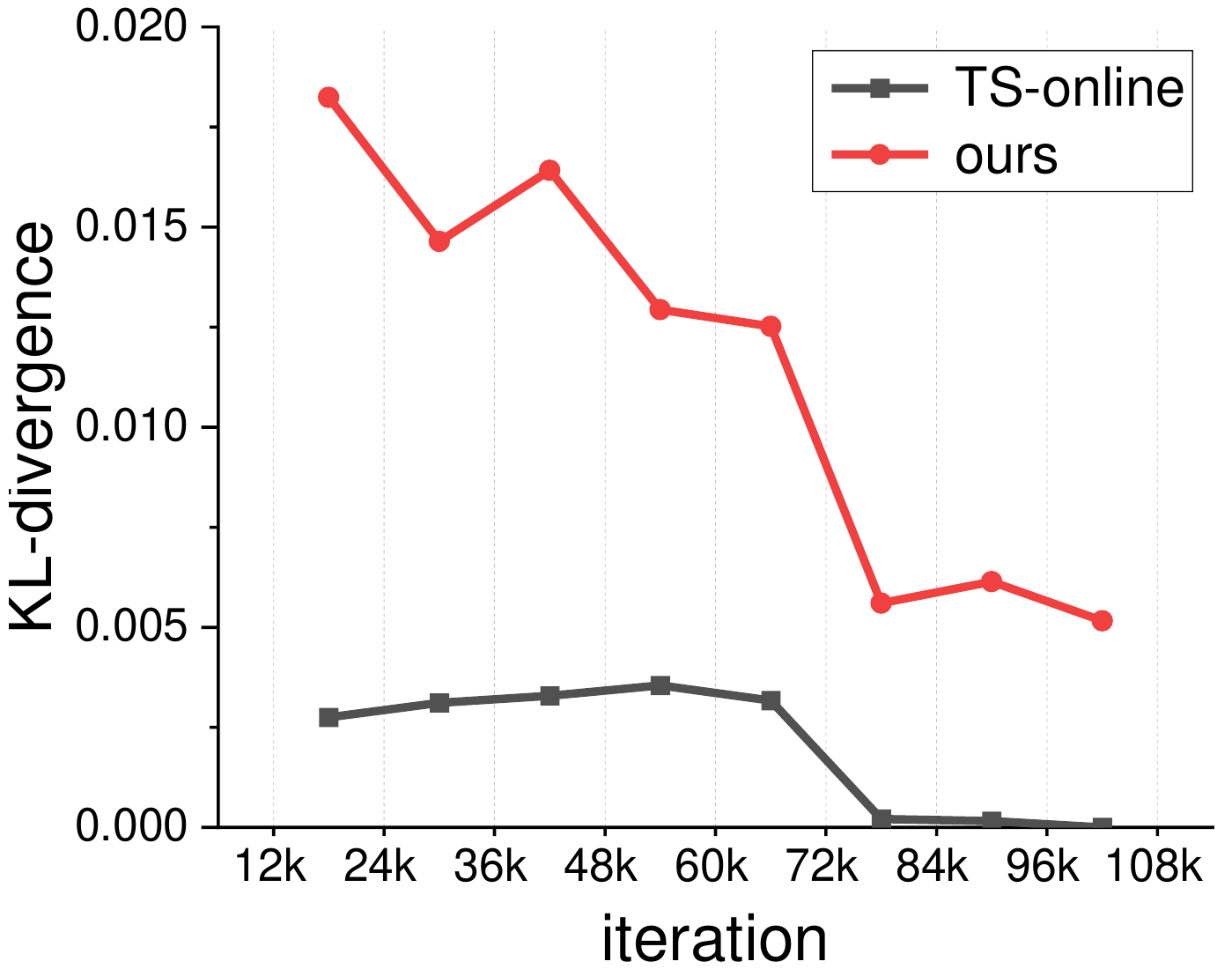}
\label{fig: kl-divergence}
\end{minipage}
\begin{minipage}[t]{0.49\linewidth}
\centering
\includegraphics[width=0.9\linewidth]{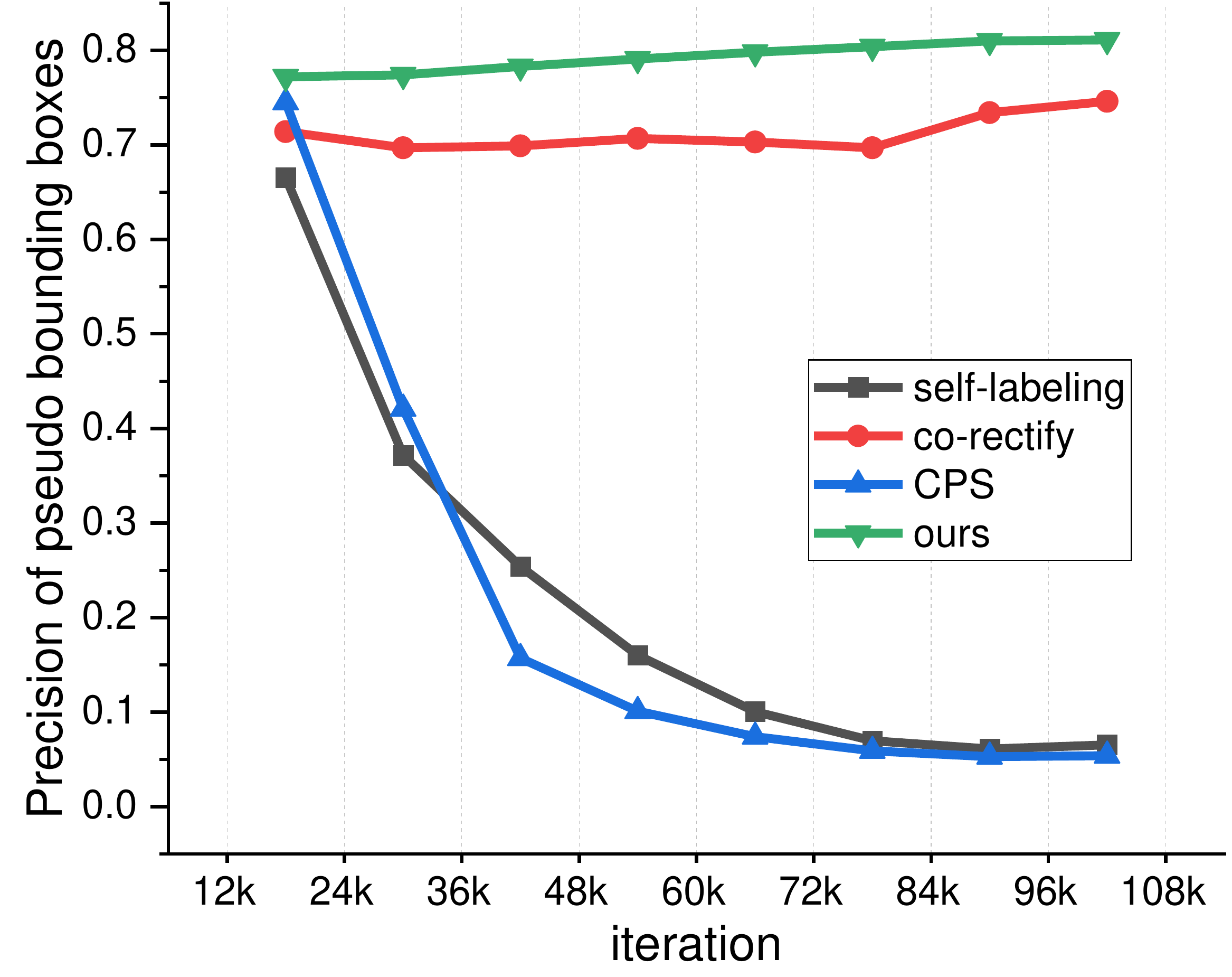}
\label{fig: pseudo label precision}
\end{minipage}
}
\vspace{-5mm}
\caption{
Left: (a) comparison on the average KL-divergence between probability scores predicted by two detectors over each 12\textit{k} iterations. ``TS-online'' stands for the teacher-student mutual learning in the online manner.
Right: (b) comparison on the average precision of pseudo bounding boxes among different methods over each 12\textit{k} iterations.
(For interpretation of the references to color in this figure legend, please refer to the online version of this article.)
}
\label{fig: discussions}
\vspace{-2mm}
\end{figure}

As for online methods~\cite{UnbiasedTeacher,HumbleTeacher,SoftTeacher}, the teacher detector is converged to the student detector and yields similar predictions in the late stage of training, thus the pseudo label generation process degenerates to the self-labeling process and suffers from the same limitations.
For instance, we conduct online teacher-student mutual learning based on SSD300 and Pascal VOC.
As shown in Fig.~\ref{fig: discussions}(a), the average KL-divergence between probability scores predicted by teacher and student detector reaches zero in the last 40\textit{k} iterations. 
Correspondingly, the detection performance shown in Table~\ref{tab: VOC results} also indicate the ineffectiveness of online teacher-student mutual learning.

\paragraph{Co-rectify and CPS}
Recently, a co-training based SSOD method named co-rectify has been proposed in~\cite{InstantTeaching}, which is the only prior work taking the idea of co-training in the SSOD task to the best of our knowledge. 
In co-rectify, the pseudo bounding boxes are first predicted by one detector, then refined by corresponding predictions from another model, with probability scores and coordinates being averaged. 
Besides, we notice that a recent work proposes cross pseudo supervision (CPS) \cite{CPS} for the semi-supervised semantic segmentation task and achieves the state-of-the-art performances, where each model directly takes the predictions from the other model as pseudo labels.
CPS can be adapted to the SSOD task, as each detector is supervised by the other detector.
However, as shown in Fig.~\ref{fig: discussions}(b), the precision values of pseudo label of these methods are inferior to that of CrossRectify (conducting semi-supervised training with SSD300 on Pascal VOC). 
%
We consider the reason behind such phenomenon lies in that simply averaging multiple predictions (co-rectify) or directly taking predictions from other models as supervision (CPS) cannot fully exploit the advantages of multiple models, in comparison to our cross-rectifying mechanism.
Their inferior performances shown in Table~\ref{tab: VOC results} and Table~\ref{tab: ablation-rectify-strategy} also validate our consideration.
Besides, we also investigate more alternative strategies on pseudo label rectification and observe that cross-rectifying turns out to be most effective strategy (as detailed in Sec.~\ref{sec: empirical study}).

\section{Experiments}
\label{sec:exp}

\subsection{Datasets and Evaluation Metrics}
\label{sec: datasets and metrics}
\paragraph{2D semi-supervised object detection}
We evaluate the proposed CrossRectify on two widely-used benchmark datasets, \ie, Pascal VOC\cite{VOC} and MS-COCO \cite{MSCOCO}.
Pascal VOC has 20 object categories. We take the VOC07 trainval set (5,011 images) as labeled and the VOC12 trainval set (11,540 images) as unlabeled. 
The detection performance is evaluated on the VOC07 test set (4,952 images) using the VOC style AP$_{50}$ metric.
MS-COCO has 80 object categories. We follow the same settings as that in \cite{STAC,UnbiasedTeacher,SoftTeacher,InstantTeaching,CombatingNoise} to randomly sample 1/2/5/10\% of the COCO2017 train set (118,287 images) as labeled and take the remaining part as unlabeled. 
Also, we create five data folds under each degree of supervision, and finally report the mean and standard deviation from five results.
The detection performance is evaluated on the COCO2017 val set (5,000 images) using the COCO style AP$_{50:95}$ metric. 

\paragraph{3D semi-supervised object detection}
We follow~\cite{3DIoUMatch} to conduct experiments on the SUN-RGBD benchmark dataset~\cite{song2015sun}.
We randomly sample 5\% of 5,285 training samples as labeled and take the remaining part as unlabeled. The detection performances is evaluated on 5,050 validation samples, using both AP$_{25}$ and AP$_{50}$ metrics.

\subsection{Implementation Details}
\label{sec: implementation details}
\paragraph{Detector structures}
We carry out experiments on the Pascal VOC dataset with two detector structures, that are SSD300~\cite{SSD} with VGG-16 backbone and Faster-RCNN-FPN~\cite{FasterRCNN,FPN} with ResNet-50 backbone.
The latter structure is also utilized in experiments on the MS-COCO dataset.
As for 3D detection, we utilize VoteNet~\cite{votenet} with PointNet++ backbone~\cite{pointnet++}.

\paragraph{Training settings}
We utilize the Pytorch implementation\footnote{\url{https://github.com/amdegroot/ssd.pytorch}} to train SSD300 on Pascal VOC. 
Within a total of 120\textit{k} iterations, we conduct fully-supervised training in the first 12\textit{k} iterations as warm-up. 
We ramp-up/down the unsupervised loss weight $\lambda_U$, and set threshold $\tau$ and batch size as 0.5 and 32 according to~\cite{ISD}.
We utilize the Detectron2 platform\footnote{\url{https://github.com/facebookresearch/detectron2}} to train Faster-RCNN-FPN on Pascal VOC.
We train a total of 36\textit{k} iterations with first 6\textit{k} being fully-supervised warm-up, and adopt the same data augmentation strategy as that in~\cite{UnbiasedTeacher}. 
We set $\lambda_U$ as 2.0 and threshold $\tau$ as 0.7 following~\cite{UnbiasedTeacher}.
The batch sizes for labeled data and unlabeled data are both 16.
The threshold $\delta$ on matching metric is 0.5. 
To show the generality across different platforms of our CrossRectify method, we adopt the MMdetection\footnote{\url{https://github.com/open-mmlab/mmdetection}} to train Faster-RCNN-FPN on MS-COCO. 
We train a total of 180\textit{k} iterations and adopt the data augmentation strategies in~\cite{SoftTeacher}. 
Under 1\% degree of supervision, we conduct fully-supervised warm-up in the first 80\textit{k} iterations to ensure the stability of training.
We set $\lambda_U$ as 4.0 and threshold $\tau$ as 0.9 according to~\cite{SoftTeacher}.
The batch sizes for labeled data and unlabeled data are respectively 8 and 32.
The threshold $\delta$ on matching metric is 0.5. 
To train VoteNet on SUN-RGBD, we first conduct fully-supervised pre-training by 900 iterations, then conduct semi-supervised training by 1\textit{k} iterations, following~\cite{3DIoUMatch}.


Note that the exponential moving average (EMA) strategy is commonly used for the pseudo label-based methods, since a detector model aggregated by EMA can yield more conservative and stable predictions than the detector itself  \cite{UnbiasedTeacher,HumbleTeacher,SoftTeacher}.
For fair comparisons, We follow such common practice in our experiments, as we utilize the EMAs of two detectors to conduct the detector feed-forward process.

\subsection{Results}
\label{sec: results}
\paragraph{Pascal VOC}Table~\ref{tab: VOC results} shows the results of our CrossRectify method compared with other training frameworks on Pascal VOC.
As for the SSD300 detector, we take the 71.73\% AP$_{50}$ performance of fully-supervised training as the baseline. 
%
As can be seen, our proposed method can obtain a 73.65\% AP$_{50}$ result, while the results of all compared approaches are only about 72.50\%.
Such comparison validates the effectiveness of our CrossRectify on improving pseudo label quality.
Besides, the WBF-merged~\cite{WBF} results from both detectors can further boost the final performances, denoted as CrossRectify$^\ast$.
Under the mix-up data augmentation~\cite{MixUp}, our CrossRectify method can still show better performance than the self-labeling-based method, ISD~\cite{ISD} (by a 1.41\% margin). 

\begin{table}[t]
\centering
\caption{2D Semi-supervised Object Detection performances (AP$_{50}$) on Pascal VOC benchmark dataset.}
\vspace{3mm}
\scalebox{.62}{
\begin{tabular}{ c | c | l | c  c | c | l }
\hline
Model & Backbone & \makecell[c]{Method} & Labeled & Unlabeled & Threshold & \makecell[c]{AP$_{50}$}
\\ \hline
\multirow{12}{*}{SSD300} & \multirow{12}{*}{VGG-16}
&  Supervised                                & VOC07&  -  &  - &71.73
\\
&& Self-Labeling                             & VOC07&VOC12& 0.5&72.13 (+0.40)
\\ 
&& Online Teacher-Student Mutual Teaching    & VOC07&VOC12& 0.5&72.56 (+0.83)
\\ 
&& Offline Teacher-Student Mutual Teaching\cite{DataUncertainty}                                                          & VOC07&VOC12& -  &72.52 (+0.79)
\\
&& Cross Pseudo Supervision\cite{CPS}        & VOC07&VOC12& -  &72.56 (+0.83)
\\ 
&& Co-rectify\cite{InstantTeaching}          & VOC07&VOC12& 0.5&72.48 (+0.75)
\\
&& CrossRectify (ours)                       & VOC07&VOC12& 0.5&73.56 (+1.83)
\\
&& CrossRectify$^\ast$ (ours)                & VOC07&VOC12& 0.5&\textbf{74.97 (+3.24)}
\\ \cdashline{3-7}
&& Supervised + MixUp                    & VOC07&VOC12& 0.5&73.04
\\
&& Self-Labeling + MixUp (ISD~\cite{ISD})       & VOC07&VOC12& 0.5&73.50 (+0.46)
\\
&& CrossRectify + MixUp  (ours)          & VOC07&VOC12& 0.5&74.91 (+1.87)
\\
&& CrossRectify$^\ast$ + MixUp (ours)    & VOC07&VOC12& 0.5&\textbf{76.16 (+3.12)}
\\\hline
\multirow{9}{*}{\shortstack[l]{Faster-\\RCNN-\\FPN}} & \multirow{9}{*}{ResNet-50}
& Supervised                     & VOC07 &   -   &  -  & 76.90
\\ 
&& CSD\cite{CSD}                 & VOC07 & VOC12 &  -  & 77.50 (+0.60)
\\ 
&& STAC\cite{STAC}               & VOC07 & VOC12 &  -  & 77.50 (+0.60)
\\ 
&& Co-rectify\cite{InstantTeaching}  
                                 & VOC07 & VOC12 &  -  & 79.20 (+2.30)
\\ 
&& Combating Noise\cite{CombatingNoise}
                                 & VOC07 & VOC12 &  -  & 80.60 (+3.70)
\\ 
&& Humble Teacher\cite{HumbleTeacher}
                                 & VOC07 & VOC12 & 0.7 & 80.94 (+3.94)
\\ 
&& Unbiased Teacher\cite{UnbiasedTeacher}
                                 & VOC07 & VOC12 & 0.7 & 80.51 (+3.61)
\\ 
&& CrossRectify (ours)           & VOC07 & VOC12 & 0.7 & 81.56 (+4.66)
\\ 
&& CrossRectify$^\ast$ (ours)    & VOC07 & VOC12 & 0.7 & \textbf{82.34 (+5.44)}
\\\hline
\end{tabular}
}
\label{tab: VOC results}
\end{table}

As for the Faster-RCNN-FPN detector, we compare CrossRectify with previous methods, and our method can improve the AP$_{50}$ result with a 4.66\% margin over fully-supervised baseline, achieving the state-of-the-art performance.
Note that Unbiased Teacher~\cite{UnbiasedTeacher} reports the performance based on the COCO style AP$_{50}$ metric in their paper.
For a fair comparison, we instead adopt the VOC style AP$_{50}$ metric for Unbiased Teacher, and the AP$_{50}$ raises from 77.37\% to 80.51\%,  still surpassed by that of CrossRectify with a 0.59\% margin.

\begin{table}[t]
\centering
\caption{2D Semi-supervised object detection performances (AP$_{50:95}$) on MS-COCO benchmark dataset.}
\vspace{2.5mm}
\scalebox{0.68}{
\begin{tabular}{c | c | c | c  c  c  c }
\hline
\multirow{2}{*}{Model} & \multirow{2}{*}{Backbone}  & \multirow{2}{*}{Method}  & \multicolumn{4}{c}{Proportion of labeled data}\\
&&&  $1\%$ & $2\%$ & $5\%$ & $10\%$
\\ \hline
\multirow{10}{*}{\shortstack[l]{Faster-\\RCNN-\\FPN}}&\multirow{10}{*}{ResNet-50}&Supervised
& 9.05 $\pm$ 0.16  & 12.70 $\pm$ 0.15  & 18.47 $\pm$ 0.22  & 23.86 $\pm$ 0.81
\\ 
&&CSD\cite{CSD}
& 10.51 $\pm$ 0.06 & 13.93 $\pm$ 0.12  & 18.63 $\pm$ 0.07  & 22.46 $\pm$ 0.08
\\ 
&&STAC\cite{STAC}                       
& 13.97 $\pm$ 0.35 & 18.25 $\pm$ 0.25  & 24.38 $\pm$ 0.12  & 28.64 $\pm$ 0.21
\\ 
&&Unbiased Teacher\cite{UnbiasedTeacher}                       
& 20.75 $\pm$ 0.12 & 24.30 $\pm$ 0.07  & 28.27 $\pm$ 0.11  & 31.50 $\pm$ 0.10
\\ 
&&Humble Teacher\cite{HumbleTeacher}
& 16.96 $\pm$ 0.38 & 21.72 $\pm$ 0.24  & 27.70 $\pm$ 0.15  & 31.61 $\pm$ 0.28
\\ 
&&Co-rectify\cite{InstantTeaching}
& 18.05 $\pm$ 0.15 & 22.45 $\pm$ 0.15  & 26.75 $\pm$ 0.05  & 30.40 $\pm$ 0.05
\\ 
&&Combating Noise\cite{CombatingNoise}
& 18.41 $\pm$ 0.10 & 24.00 $\pm$ 0.15  & 28.96 $\pm$ 0.29  & 32.43 $\pm$ 0.20
\\ 
&&Soft Teacher\cite{SoftTeacher}
& 20.46 $\pm$ 0.39 & 26.20 $\pm$ 0.10  & 30.74 $\pm$ 0.08  & 34.04 $\pm$ 0.14
\\ 
&&CrossRectify (ours)                 
& 21.90 $\pm$ 0.11
& 26.70 $\pm$ 0.07
& 31.70 $\pm$ 0.04
& 34.89 $\pm$ 0.07
\\ 
&&CrossRectify$^\ast$ (ours)                 
& \textbf{22.50 $\pm$ 0.12}
& \textbf{27.60 $\pm$ 0.07}
& \textbf{32.80 $\pm$ 0.05}
& \textbf{36.30 $\pm$ 0.07}
\\\hline
\end{tabular}
}
\label{tab: MS-COCO results}
\end{table}
\begin{figure*}[t]
\centering
{
\includegraphics[width=0.9\linewidth]{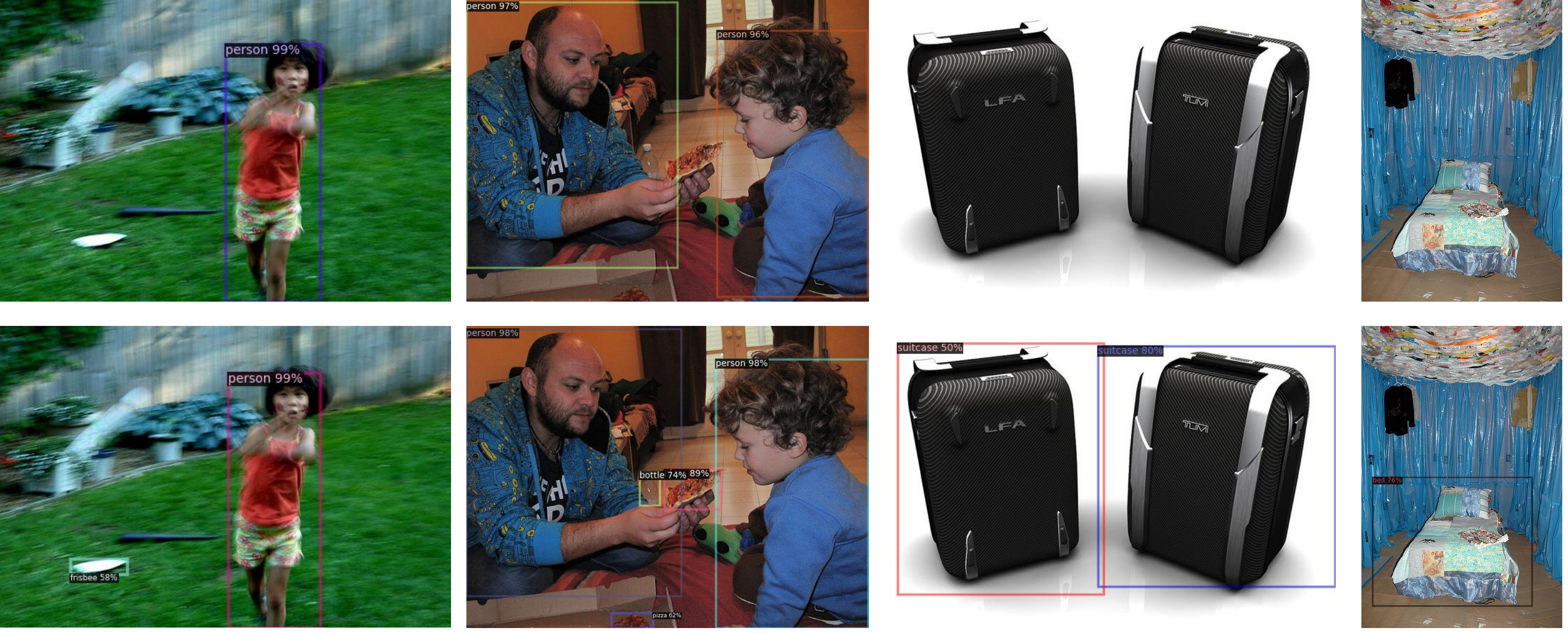}
}
\caption{The visual comparisons between Self-Labeling (the first row) and the proposed CrossRectify method (the second row) on MS-COCO under 1\% degree of supervision.  
(For color discrimination in this figure, please refer to the online version of this article.)}
\label{fig: visualization}
\end{figure*}

\paragraph{MS-COCO}
Table~\ref{tab: MS-COCO results} shows the performances of our CrossRectify method compared with previous state-of-the-arts on the MS-COCO dataset.
Under different degrees of supervision 1\%, 2\%, 5\% and 10\%, the proposed CrossRectify can obtain consistent and substantial improvements, surpassing those of Soft Teacher~\cite{SoftTeacher} by 1.46\%, 0.50\%, 0.96\%, and 0.85\% AP$_{50:95}$ margins.
These comparative results further verify the effectiveness of the proposed method.
Moreover, we visualize the pseudo bounding boxes for some unlabeled images in Fig.~\ref{fig: visualization}. Compared with self-labeling training scheme, our method can yield more accurate pseudo boxes.

\paragraph{SUN-RGBD} Table~\ref{tab: SUN-RGBD results} shows the comparison with all previous works (\ie, SESS~\cite{SESS} and 3DIoUMatch~\cite{3DIoUMatch}) on the SUN-RGBD benchmark dataset.
Under 5\% degree of supervision, the performance of our CrossRectify can outperform that of the state-of-the-art 3DIoUMatch method by 3.1 AP$_{25}$ and 1.9 AP$_{50}$ margins.
The results validate the efficiency of CrossRectify on 3D semi-supervised object detection task.
We omit the WBF-merging performance CrossRectify$^\ast$, because WBF does not support on 3D bounding boxes with different rotation angles.

\begin{table}[t]
\centering
\caption{3D Semi-supervised object detection performances (AP$_{25}$ and AP$_{50}$) on SUN-RGBD benchmark dataset.}
\vspace{2.5mm}
\scalebox{.79}{
\begin{tabular}{ c | c | c | c c }
\hline
Model & Backbone & Method & AP$_{25}$ & AP$_{50}$
\\ \hline
\multirow{4}{*}{VoteNet} & \multirow{4}{*}{PointNet++}
& Supervised                  & 29.9 $\pm$ 1.5 & 10.5 $\pm$ 0.5 \\
&&SESS\cite{SESS}             & 34.2 $\pm$ 2.0 & 13.1 $\pm$ 1.0 \\
&&3DIoUMatch\cite{3DIoUMatch} & 39.0 $\pm$ 1.9 & 21.1 $\pm$ 1.7 \\
&&CrossRectify (ours)         & \textbf{42.1 $\pm$ 1.7} & \textbf{23.0 $\pm$ 1.2} \\\hline
\end{tabular}
}
\label{tab: SUN-RGBD results}
\end{table}

\subsection{Empirical Study}
\label{sec: empirical study}
\paragraph{Pseudo label rectification strategy}
Now we investigate the alternative strategies on rectifying the pseudo labels, including
(a) only utilizing the intersection of two prediction sets from two detectors as pseudo labels, which is composed of the objects classified as the same classes.
(b) Only utilizing the difference set of two prediction sets from two detectors as the pseudo label, which is composed of objects classified as different classes.
(c) Directly taking all the predicted bounding boxes from the other detector as the pseudo label, which turns out to be the cross pseudo supervision (CPS)~\cite{CPS}. 
As observed in Table.~\ref{tab: ablation-rectify-strategy}, all these strategies cannot ensure the pseudo label quality and finally lead to inferior performances.

\paragraph{Extension to more detectors}
Our proposed CrossRectify method can be easily extended to train more than two detectors simultaneously. 
Specifically, during the pseudo label rectification process, each pseudo bounding box is re-labeled by the majority of all predicted classes, and re-located by the average of all predicted coordinates.
As displayed in Table~\ref{tab: ablation_four_detectors}, CrossRectify over four SSD300 detectors can bring a 0.06\% AP$_{50}$ improvement for each detector on average. 
We believe that two-detector scenario is already able to cross-rectify the misclassified pseudo labels adequately. 

\begin{table}[t]
\centering
\caption{Empirical study on strategies of pseudo label rectification.}
\vspace{3mm}
\scalebox{0.7}{
\begin{tabular}{c | c c | c | c}
\hline
Model                   & Labeled & Unlabeled & Strategy         & AP$_{50}$
\\ \hline
\multirow{5}{*}{SSD300} & VOC07   &   -       &  -               & 71.73
\\ 
                        & VOC07   & VOC12     & intersection     & 72.59
\\
                        & VOC07   & VOC12     & difference set   & 65.52
\\                        
                        & VOC07   & VOC12     & CPS              & 72.56
\\ 
                        & VOC07   & VOC12     & CrossRectify     &\textbf{73.65}
\\\hline
\end{tabular}
}
\label{tab: ablation-rectify-strategy}
\end{table}
\begin{table}[t]
\centering
\caption{Extension to four detector models on Pascal VOC dataset.}
\vspace{3mm}
\scalebox{.7}{
\begin{tabular}{ c c c c c c}
\hline
Model & Index & Single & Average & WBF-Merged
\\ \hline
\multirow{6}{*}{SSD300}  & detector \#1     &   73.67   &   \multirow{2}{*}{73.65}  &   \multirow{2}{*}{74.83}
\\ 
&detector \#2     &   73.63   &                           &
\\ \cline{2-5}
&detector \#1     &   73.60   &   \multirow{4}{*}{73.71}  &   \multirow{4}{*}{75.84}
\\ 
&detector \#2     &   73.71   &                           &
\\ 
&detector \#3     &   73.80   &                           &
\\ 
&detector \#4     &   73.73   &                           &
\\\hline
\end{tabular}
}
\vspace{-0mm}
\label{tab: ablation_four_detectors}
\vspace{-0mm}
\end{table}

\section{Conclusion}
\label{sec: conclusion}
In this paper, we propose the CrossRectify training framework for the semi-supervised object detection task, aiming to address the inherent limitations in the self-labeling-based methods.
In CrossRectify, two detectors with same structure but different initialization are trained simultaneously. The disagreements on the same objects between two detectors are utilized to discern and rectify the latent self-errors predicted by each single detector.
Moreover, we conduct both theoretical analysis and quantitative experiments to show the superiority on improving pseudo label quality, compared with other recent works. 
Extensive results on both 2D and 3D semi-supervised object detection task validate the effectiveness and versatility of CrossRectify.

\section*{Declaration of Competing Interest}
The authors declare that they have no known competing finan- cial interests or personal relationships that could have appeared to influence the work reported in this paper.

\section*{Acknowledgement}
This work was supported by the National Natural Science Foundation of China 61832016, U20B2070, 6210070958, 62102162.

\bibliography{mybibfile}

\end{document}